\documentclass[letterpaper]{article} 
\usepackage[draft]{aaai2027}  
\usepackage[hyphens]{url}  
\usepackage{graphicx} 
\usepackage{natbib}  
\usepackage{caption} 
\usepackage{algorithm}
\usepackage{algorithmic}
\usepackage{enumitem}
\usepackage{amsmath}
\usepackage{amssymb}
\usepackage{booktabs}
\usepackage{tabularx}
\usepackage{array}
\usepackage{makecell}
\usepackage{multirow}
\usepackage{graphicx}
\usepackage[table]{xcolor}
\definecolor{sectiongray}{RGB}{242,242,242}
\definecolor{oursblue}{RGB}{232,244,248}
\definecolor{selectedred}{RGB}{149,55,52}

\definecolor{codeblue}{RGB}{48,88,145}
\definecolor{codeurl}{RGB}{45,82,135}
\newcolumntype{Y}{>{\centering\arraybackslash}X}

\usepackage{newfloat}
\usepackage{listings}
\DeclareCaptionStyle{ruled}{labelfont=normalfont,labelsep=colon,strut=off} 
\floatstyle{ruled}
\newfloat{listing}{tb}{lst}{}
\floatname{listing}{Listing}

\usepackage{booktabs}

\title{InsertFuse: A Unified Framework for Multi-Category Reference-Guided Image Insertion}
\author{
    Guangzhao Li\textsuperscript{\rm 1,2}\equalcontrib,
    Qingyan Wei\textsuperscript{\rm 1}\equalcontrib,
    Huayu Zheng\textsuperscript{\rm 1},
    Yige Zheng\textsuperscript{\rm 1},
    Chaoyang Zhang\textsuperscript{\rm 2,4},
    Jie Yang\textsuperscript{\rm 2,5},
    Yunan Ding\textsuperscript{\rm 3},
    Yan Tai\textsuperscript{\rm 1},
    Siqi Luo\textsuperscript{\rm 1},
    Xiaohong Liu\textsuperscript{\rm 1,2}\corresponding
}

\affiliations{
    \textsuperscript{\rm 1}Shanghai Jiao Tong University
    \quad
    \textsuperscript{\rm 2}Shanghai Innovation Institute
    \quad
    \textsuperscript{\rm 3}The Hong Kong Polytechnic University \\
    \quad
    \textsuperscript{\rm 4}Xi'an Jiaotong University
    \quad
    \textsuperscript{\rm 5}Wuhan University
    \\[4pt]

    \textcolor{codeblue}{\textbf{Homepage:}}
    \url{https://insertfuse.github.io}
}

\begin{document}

\maketitle

\begin{abstract}
We present \textbf{InsertFuse}, a unified framework for multi-category
reference-guided image insertion.
Its key idea is to decouple category-specific expertise learning from
cross-category capability consolidation.
InsertFuse first trains specialized experts for different insertion
categories and then introduces \textbf{Insertion On-Policy Distillation
(IOPD)} to consolidate their capabilities into a single student.
By querying the matched expert at states visited by the student, IOPD
preserves category-specific insertion behavior while mitigating the
cross-category interference caused by direct joint training.
To improve spatial control, we propose \textbf{Token-Aligned Geometry
Conditioning (TAGC)}, which maps mask-derived geometric cues to the
visual token grid, and \textbf{Region-Balanced Flow Matching}, which
separately normalizes prediction errors inside and outside the insertion
region to prevent background-dominated and scale-dependent supervision.
We further introduce \textbf{Reference CFG} to isolate and strengthen
the guidance induced by the visual reference under fixed scene and
geometry conditions, with IOPD transferring this enhanced supervision
into the unified student.
Extensive experiments on the public AnyInsertion benchmark and our
multi-category test set demonstrate state-of-the-art performance on most
metrics, showing strong reference fidelity and
generation quality across diverse insertion categories.
\end{abstract}

\section{Introduction}

Recent advances in diffusion models and flow matching~\cite{ho2020ddpm,rombach2022ldm,peebles2023dit,lipman2023flowmatching,esser2024sd3} have significantly improved image generation~\cite{zheng2024cogview3, li2024hunyuandit, xie2024sana, qin2025luminaimage2, imagen3_2024, cai2025hidream, wu2025qwenimage, labs2025flux1kontextflowmatching} and editing~\cite{zhang2025icedit, liu2025step1xedit, chen2024unireal, yu2025anyedit} in terms of visual quality, semantic understanding, and complex instruction following. Building on this progress, using reference images to achieve more precise control over generated content has attracted increasing attention~\cite{chen2024anydoor, song2025insertanything, zheng2026a2edit}. Reference-guided image insertion is a representative task in this setting that aims to place a target subject from a reference image into a designated region, while preserving its appearance and identity and adapting its scale and spatial placement to the new scene. This task has broad applications in virtual try-on, product presentation, and interactive content creation.

Despite recent progress in image insertion, building a unified model that can handle diverse insertion tasks remains challenging. Although different insertion categories share similar input formats, they require substantially different generation strategies and modeling objectives. Accessory insertion often demands fine-grained texture preservation~\cite{chen2024anydoor,winter2025objectmate}, virtual try-on requires complex non-rigid deformation, and human insertion places greater emphasis on identity consistency~\cite{wang2024instantid,li2023photomaker,
song2025insertanything}. Directly training these heterogeneous tasks within a single model can introduce conflicting optimization signals, thereby substantially weakening the specialized capabilities required by each task. Recent work~\cite{zheng2026a2edit} alleviates this
conflict through a Mixture of Transformers, which dynamically
routes input features to specialized expert branches.
However, since these experts are still jointly optimized within a
unified training process, the conflicts among category-specific
modeling objectives are only partially alleviated rather than
fundamentally resolved.

To address these challenges, we propose \textbf{InsertFuse}, a unified framework for multi-category reference-guided image insertion. Its core idea is to separate the learning of category-specific expertise from the consolidation of cross-category capabilities. Specifically, we introduce \textbf{Insertion On-Policy Distillation (IOPD)}, an on-policy distillation framework designed for image insertion. In the first stage, we train separate experts for different insertion categories, allowing each expert to acquire specialized knowledge from its corresponding data distribution. In the second stage, we employ on-policy distillation to consolidate their category-specific expertise into a unified student model. This two-stage design preserves task-specific capabilities while alleviating the cross-task interference caused by direct joint training.
However, resolving cross-category interference alone is insufficient for high-quality reference-guided insertion. Across different insertion categories, the model must satisfy two fundamental requirements: it should accurately confine the generated content to the designated region while preserving the surrounding scene, and it should retain the distinctive appearance and identity of the reference subject after adapting it to the target context.

Existing formulations~\cite{song2025insertanything, zheng2026a2edit} provide limited support for these requirements.
Simply concatenating a binary mask with the model input offers only
coarse spatial guidance, while a globally averaged flow-matching
objective can be dominated by the substantially larger background
region.
To provide more explicit and balanced spatial supervision, we introduce
\textbf{Token-Aligned Geometry Conditioning (TAGC)}, which maps the
insertion mask into position-dependent geometric conditions aligned
with the target token grid.
We further propose \textbf{Region-Balanced Flow Matching}, which
separately normalizes the prediction errors inside and outside the
insertion region, preventing the supervision strength from being
dominated by the background area or varying with the insertion scale.
Together, these designs provide more effective spatial guidance and
balanced regional optimization, improving insertion accuracy while
preserving the surrounding scene.

Reference fidelity presents a complementary challenge. Conventional classifier-free guidance~\cite{ho2022classifier0free} in image editing is commonly constructed by dropping only the text condition, such that the resulting guidance signal does not explicitly isolate the contribution of the visual reference. We therefore introduce \textbf{Reference CFG}, which constructs a reference-free prediction while retaining the source scene and spatial conditions, thereby providing dedicated guidance for preserving the appearance and identity of the reference subject. Through IOPD, these reference-enhanced expert policies are further distilled into a single student, retaining strong reference consistency without requiring an additional Reference CFG branch during inference.

We conduct extensive experiments across multiple datasets. The results show that InsertFuse substantially outperforms existing reference-guided image insertion methods, as well as several unified image editing models, in reference fidelity, spatial controllability, and generation quality. In summary, our main contributions are as follows:

\begin{itemize}

    \item We propose InsertFuse, a unified multi-category reference-guided insertion framework that decouples category-specific expert learning from cross-category consolidation via IOPD to mitigate interference from direct joint training.

    \item We introduce TAGC and Region-Balanced Flow Matching to inject token-aligned mask geometry and rebalance insertion and background supervision, improving spatial alignment, local detail, and background preservation.

    \item We propose Reference CFG to strengthen visual-reference guidance under fixed scene conditions.

    \item Extensive experiments demonstrate that InsertFuse delivers state-of-the-art performance across multiple metrics and significantly outperforms existing methods.

\end{itemize}

\section{Related Work}

\subsubsection{Image Editing.}
Early diffusion-based methods achieve localized edits through
noise-and-denoise inversion~\cite{meng2021sdedit}, cross-attention
control~\cite{hertz2022prompt}, or prompt-difference
masks~\cite{couairon2022diffedit}.
Instruction-driven editing was advanced by
InstructPix2Pix~\cite{brooks2023instructpix2pix}, the human-annotated
MagicBrush~\cite{zhang2023magicbrush}, and large-scale synthetic datasets
such as UltraEdit~\cite{zhao2024ultraedit} and
AnyEdit~\cite{yu2025anyedit}.
Unified models further broaden task coverage through multi-task
embeddings in Emu Edit~\cite{sheynin2023emuedit}, joint
generation--editing architectures in OmniGen~\cite{xiao2024omnigen} and
Qwen-Image~\cite{wu2025qwenimage}, and shared conditioning in
ACE++~\cite{mao2025aceplusplus}.
However, they are not specifically designed for the precise regional
control, reference fidelity, and background preservation required by
reference-guided insertion.

\subsubsection{Image Insertion.}
Reference-guided image insertion places a reference subject into a
designated region while preserving its appearance and adapting it to the
target scene.
Paint-by-Example~\cite{yang2023paintbyexample} conditions generation on
reference crops, whereas ObjectStitch~\cite{song2023objectstitch} injects
object features into pretrained inpainting models.
AnyDoor~\cite{chen2024anydoor} jointly models high-level identity and
low-level details for zero-shot insertion; subsequent methods expand
supervision and task scope through multi-view cross-scene
observations~\cite{winter2025objectmate} and joint removal--insertion
modeling~\cite{yu2025omnipaint}, or improve spatial control using
geometric and depth conditions~\cite{li2024bifrost} and interactive 3D
proxies~\cite{gong2026direct}.
Closely related virtual try-on methods, including
StableVITON~\cite{kim2024stableviton},
IDM-VTON~\cite{choi2024idmvton}, and
CatVTON~\cite{chong2024catvton}, focus on non-rigid garment transfer and
identity preservation.

For multi-category insertion, Insert Anything~\cite{song2025insertanything}
uses a shared DiT for human, object, and garment insertion, but directly
optimizes mixed-category data and relies largely on implicit spatial and
reference conditioning.
A$^2$-Edit~\cite{zheng2026a2edit} mitigates category interference through
Mixture-of-Transformers routing; however, its expert branches remain
jointly optimized, leaving conflicts among category-specific modeling
objectives only partially resolved.
In contrast, InsertFuse decouples specialization from consolidation by
training category-specific experts and distilling them into a
single-branch student through IOPD, together with explicit spatial and
reference guidance.
\section{Method}
\label{sec:method}

\begin{figure*}[t]
    \centering
    \includegraphics[width=0.95\textwidth]{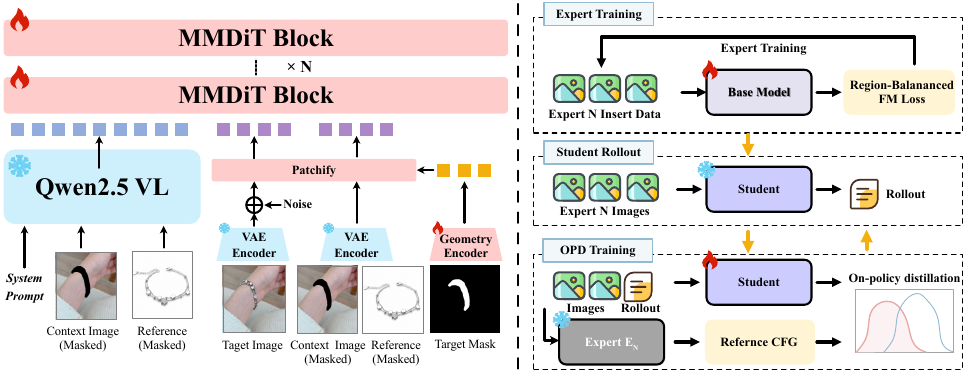}
    \caption{
        \textbf{Overview of InsertFuse.}
        \textbf{Left:} The masked context and reference images are encoded
        through the vision-language and VAE branches. Token-Aligned Geometry
        Conditioning (TAGC) converts the target mask into grid-aligned geometry
        features and integrates them with the patchified visual features.
        \textbf{Right:} Category-specific experts are first trained on their respective insertion data.
During student training, on-policy rollouts and Insertion On-Policy Distillation (IOPD) are interleaved: the student generates trajectories, and the matched frozen expert supplies velocity targets at the visited states for distillation.
The resulting unified student handles all insertion categories without retaining the experts at inference.
    }
    \label{fig:method_overview}
\end{figure*}

\subsection{Task Formulation and Framework Overview}
\label{sec:method_overview}

\noindent\textbf{Task Formulation.}
Given a complete target image \(\mathbf{I}^{\mathrm{gt}}\), a reference image \(\mathbf{I}^{\mathrm{ref}}\), and their corresponding target and reference masks \(\mathbf{M}\) and \(\mathbf{M}^{\mathrm{ref}}\), we first construct the reference subject and the masked source scene as:
\begin{equation}
    \overline{\mathbf{I}}^{\mathrm{ref}}
    =
    \mathbf{I}^{\mathrm{ref}}
    \odot
    \mathbf{M}^{\mathrm{ref}},
    \qquad
    \mathbf{I}^{\mathrm{ctx}}
    =
    \mathbf{I}^{\mathrm{gt}}
    \odot
    \left(
        1-\mathbf{M}
    \right).
    \label{eq:insertion_inputs}
\end{equation}
We then encode them using the VAE encoder \(\mathcal{V}_{\mathrm{enc}}\) of the pretrained image-editing model:
\begin{equation}
    \mathbf{z}_0
    =
    \mathcal{V}_{\mathrm{enc}}
    \left(
        \mathbf{I}^{\mathrm{gt}}
    \right),
    \,
    \mathbf{z}^{\mathrm{ctx}}
    =
    \mathcal{V}_{\mathrm{enc}}
    \left(
        \mathbf{I}^{\mathrm{ctx}}
    \right),
    \,
    \mathbf{z}^{\mathrm{ref}}
    =
    \mathcal{V}_{\mathrm{enc}}
    \left(
        \overline{\mathbf{I}}^{\mathrm{ref}}
    \right),
    \label{eq:latent_inputs}
\end{equation}
where \(\mathbf{z}_0\) is the clean target latent, \(\mathbf{z}^{\mathrm{ctx}}\) is the source-context latent, and \(\mathbf{z}^{\mathrm{ref}}\) is the reference-subject latent.

\medskip

\noindent\textbf{Flow-Matching Formulation.}
We build InsertFuse upon a pretrained flow-matching image-editing model.
Given Gaussian noise
\(
\boldsymbol{\epsilon}\sim\mathcal{N}(\mathbf{0},\mathbf{I})
\)
and a noise level \(\sigma_t\in[0,1]\), the interpolated latent and its target velocity are defined as:
\begin{equation}
    \mathbf{z}_t
    =
    (1-\sigma_t)\mathbf{z}_0
    +
    \sigma_t\boldsymbol{\epsilon},
    \qquad
    \mathbf{u}_t
    =
    \boldsymbol{\epsilon}
    -
    \mathbf{z}_0.
    \label{eq:flow_matching_path}
\end{equation}
Conditioned on the source-context and reference-subject latents, the model predicts
\(
\mathbf{v}_{\theta}
\bigl(
\widetilde{\mathbf{z}}_t,
t;
\widetilde{\mathbf{z}}^{\mathrm{ctx}},
\mathbf{z}^{\mathrm{ref}}
\bigr)
\),
where \(\widetilde{\mathbf{z}}_t\) and
\(\widetilde{\mathbf{z}}^{\mathrm{ctx}}\) denote the target and source-context latents augmented with the proposed
token-aligned geometry condition, as detailed in
Sec.~\ref{sec:tagc}.
The standard flow-matching objective is:
\begin{equation}
    \mathcal{L}_{\mathrm{FM}}
    =
    \mathbb{E}_{\mathbf{z}_0,t,\boldsymbol{\epsilon}}
    \left[
        \left\|
            \mathbf{v}_{\theta}
            \left(
                \widetilde{\mathbf{z}}_t,
                t;
                \widetilde{\mathbf{z}}^{\mathrm{ctx}},
                \mathbf{z}^{\mathrm{ref}}
            \right)
            -
            \mathbf{u}_t
        \right\|_2^2
    \right],
    \label{eq:standard_flow_matching}
\end{equation}

\begin{table*}[t]
    \centering

    \small
    \setlength{\tabcolsep}{3.5pt}
    \renewcommand{\arraystretch}{1.08}

    \resizebox{\textwidth}{!}{%
    \begin{tabular}{lcccccccccccc}
        \toprule

        \multirow{2}{*}{\textbf{Method}}
        & \multicolumn{6}{c}{\textbf{AnyInsertion}}
        & \multicolumn{6}{c}{\textbf{Our Multi-Category Test Set}} \\

        \cmidrule(lr){2-7}
        \cmidrule(lr){8-13}

        & \textbf{DINO-I}$\uparrow$
        & \textbf{CLIP-I}$\uparrow$
        & \textbf{PSNR}$\uparrow$
        & \textbf{SSIM}$\uparrow$
        & \textbf{LPIPS}$\downarrow$
        & \textbf{FID}$\downarrow$
        & \textbf{DINO-I}$\uparrow$
        & \textbf{CLIP-I}$\uparrow$
        & \textbf{PSNR}$\uparrow$
        & \textbf{SSIM}$\uparrow$
        & \textbf{LPIPS}$\downarrow$
        & \textbf{FID}$\downarrow$ \\

        \midrule
        \multicolumn{13}{c}{
            \textbf{\textit{(a) Reference-Guided Insertion Methods}}
        } \\
        \cmidrule(lr){1-13}

        AnyDoor~\cite{chen2024anydoor}
        & 0.6939
        & 0.8544
        & 17.77
        & 0.6902
        & 0.2553
        & 4.75
        & 0.6331
        & 0.8769
        & 18.57
        & 0.7606
        & 0.1999
        & 1.98 \\

        Insert Anything~\cite{song2025insertanything}
        & \textbf{0.7884}
        & \underline{0.9070}
        & \underline{23.83}
        & \underline{0.8687}
        & \textbf{0.0855}
        & \underline{1.55}
        & \underline{0.7009}
        & \underline{0.8996}
        & \underline{24.35}
        & \textbf{0.8997}
        & \underline{0.0825}
        & \textbf{1.35} \\

        A$^2$-Edit~\cite{zheng2026a2edit}
        & 0.7586
        & 0.8965
        & 22.25
        & 0.8427
        & 0.1127
        & 2.31
        & 0.6836
        & 0.8926
        & 23.56
        & 0.8930
        & 0.0924
        & 1.48 \\

        \midrule
        \multicolumn{13}{c}{
            \textbf{\textit{(b) General-Purpose Image Editing Models}}
        } \\
        \cmidrule(lr){1-13}

        FLUX.1 Kontext~\cite{labs2025flux1kontextflowmatching}
        & 0.7219
        & 0.8771
        & 20.58
        & 0.8366
        & 0.1339
        & 3.87
        & 0.5170
        & 0.8525
        & 20.45
        & 0.8820
        & 0.1126
        & 1.84 \\

        Qwen-Image-Edit-2511~\cite{wu2025qwenimage}
        & 0.7705
        & 0.8987
        & 21.73
        & 0.8090
        & 0.1158
        & 1.82
        & 0.6164
        & 0.8670
        & 19.16
        & 0.7492
        & 0.1814
        & 1.43 \\

        Qwen-Image-2.0~\cite{zhao2026qwenimage20technicalreport}
        & 0.7671
        & 0.8954
        & 21.66
        & 0.8074
        & 0.1175
        & 1.98
        & 0.6110
        & 0.8659
        & 18.95
        & 0.7428
        & 0.1863
        & 2.51 \\

        \midrule
        \multicolumn{13}{c}{
            \textbf{\textit{(c) Our Method}}
        } \\
        \cmidrule(lr){1-13}

        \textbf{InsertFuse}
        & \underline{0.7838}
        & \textbf{0.9156}
        & \textbf{24.26}
        & \textbf{0.8850}
        & \underline{0.0948}
        & \textbf{1.49}
        & \textbf{0.7148}
        & \textbf{0.9020}
        & \textbf{24.40}
        & \underline{0.8967}
        & \textbf{0.0795}
        & \underline{1.39} \\

        \bottomrule
    \end{tabular}%
    }

    \caption{
        \textbf{Quantitative comparison on the public AnyInsertion
        benchmark and our multi-category test set.}
        The best and second-best results in each column are highlighted
        in \textbf{bold} and \underline{underlined}, respectively.
    }
    \label{tab:main_comparison}
\end{table*}

\begin{figure*}[t]
    \centering
    \includegraphics[
        width=0.95\textwidth
    ]{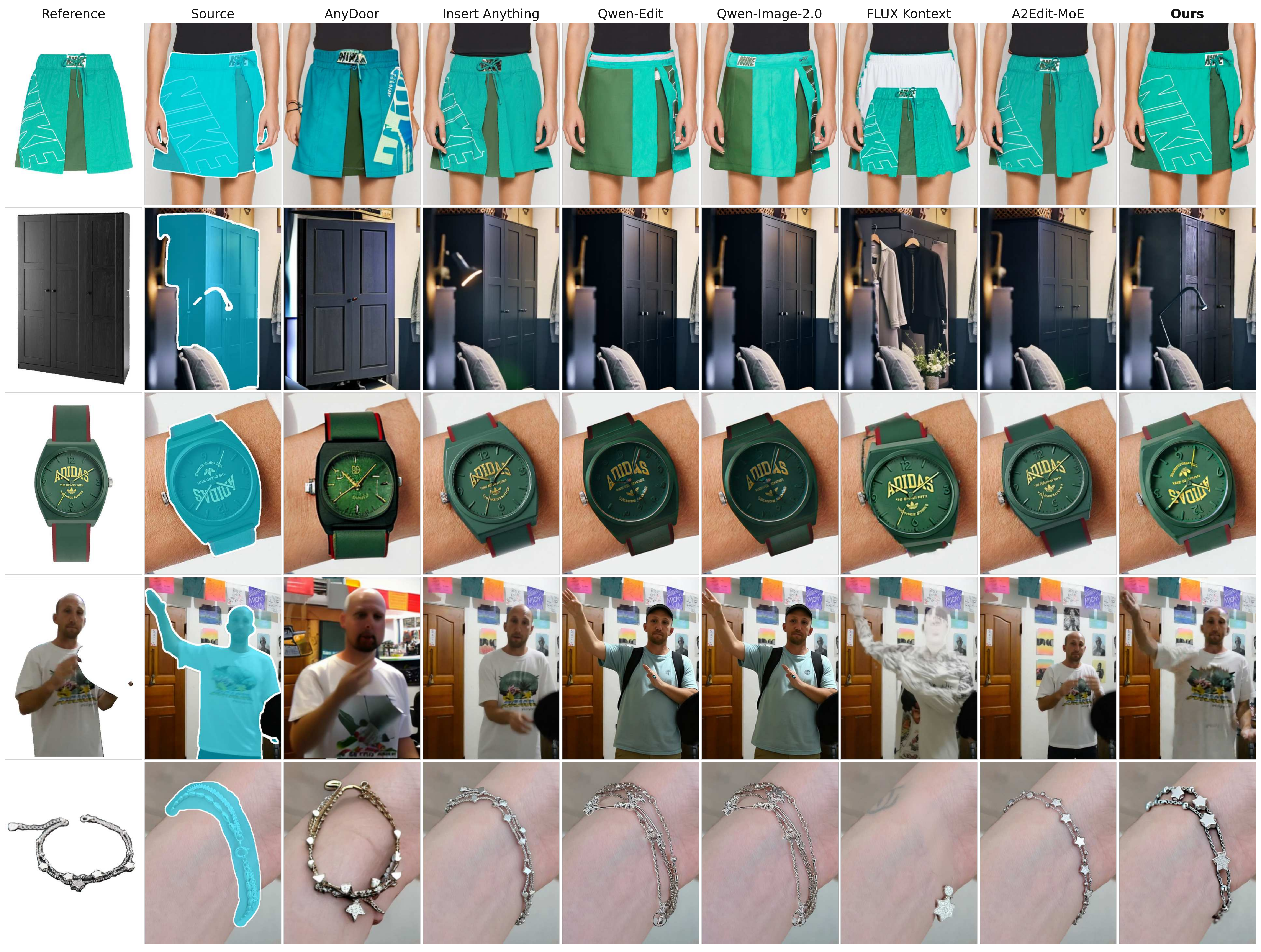}
    \caption{
        \textbf{Qualitative comparison across diverse reference-guided
        insertion cases.}
        The cyan overlay in each source image denotes the designated
        insertion region.
        InsertFuse more faithfully preserves reference-specific
        appearance and fine-grained structure while adapting the
        inserted subject to the target geometry, pose, perspective,
        and occlusion relationships.
    }
    \label{fig:qualitative_comparison}
\end{figure*}

\medskip

\noindent\textbf{Framework Overview.}
As illustrated in Fig.~\ref{fig:method_overview}, InsertFuse separates category-specific policy acquisition from cross-category capability consolidation through a two-stage framework.
In the first stage, we therefore initialize multiple category-specific experts from the same pretrained image-editing model and train each expert on its corresponding insertion category.
We further introduce Token-Aligned Geometry Conditioning (TAGC), Region-Balanced Flow Matching, and Reference CFG to equip these experts with accurate spatial control, balanced regional optimization, and strong reference fidelity.
In the second stage, we initialize a unified student from the same pretrained model, freeze all category-specific experts, and consolidate their specialized capabilities through Insertion On-Policy Distillation.

\subsection{Category-Specific Expert Learning}
\label{sec:expert_learning}

We train one expert per category using the following shared insertion-oriented components.

\subsubsection{Token-Aligned Geometry Conditioning. }
\label{sec:tagc}
Existing insertion methods commonly inject the insertion mask by concatenating it with the visual inputs and encoding them jointly. While compatible with pretrained image-editing backbones, this formulation introduces additional conditioning overhead and requires the model to implicitly learn the correspondence between the mask and the visual content. 

We instead introduce Token-Aligned Geometry Conditioning (TAGC), which encodes the insertion mask into geometry-aware embeddings and injects them directly into both the target and source-context branches.
Given the insertion mask \(\mathbf{M}\), we first construct a compact
geometric representation:
\begin{equation}
    \mathbf{G}
    \left(
        \mathbf{M}
    \right)
    =
    \left[
        \mathbf{M},
        \tanh
        \left(
            \frac{D_{\mathbf{M}}}{\tau}
        \right),
        \exp
        \left(
            -
            \frac{
                \left|D_{\mathbf{M}}\right|
            }{\tau}
        \right)
    \right],
    \label{eq:mask_geometry}
\end{equation}
where \(D_{\mathbf{M}}\) denotes the signed distance transform of the mask
and \(\tau\) controls the spatial scale.
This representation preserves both the regional occupancy and
boundary-aware geometry of the insertion area.
A lightweight geometry encoder \(\Phi_{\mathrm{geo}}\) extracts a shared
geometry feature:
\begin{equation}
    \mathbf{e}^{\mathrm{geo}}
    =
    \Phi_{\mathrm{geo}}
    \left(
        \mathbf{G}(\mathbf{M})
    \right).
    \label{eq:shared_geo_feature}
\end{equation}
Since the target and source-context branches serve different roles, we
further employ two branch-specific projections:
\begin{equation}
    \Delta \mathbf{z}^{\mathrm{tar}}
    =
    P_{\mathrm{tar}}
    \left(
        \mathbf{e}^{\mathrm{geo}}
    \right),
    \qquad
    \Delta \mathbf{z}^{\mathrm{ctx}}
    =
    P_{\mathrm{ctx}}
    \left(
        \mathbf{e}^{\mathrm{geo}}
    \right).
    \label{eq:dual_geo_projection}
\end{equation}
The resulting geometry embeddings are injected into the two branches by
residual addition:
\begin{equation}
    \widetilde{\mathbf{z}}_{t}
    =
    \mathbf{z}_{t}
    +
    \Delta \mathbf{z}^{\mathrm{tar}},
    \qquad
    \widetilde{\mathbf{z}}^{\mathrm{ctx}}
    =
    \mathbf{z}^{\mathrm{ctx}}
    +
    \Delta \mathbf{z}^{\mathrm{ctx}}.
    \label{eq:dual_tagc_injection}
\end{equation}
The geometry-enhanced latents
\(\widetilde{\mathbf{z}}_{t}\) and
\(\widetilde{\mathbf{z}}^{\mathrm{ctx}}\) are then used as the target and source-context inputs to the velocity predictor. Injecting geometry into the target branch explicitly indicates where content should be generated or preserved, while its injection into the context branch identifies the missing region and its boundary in the source scene. 

This dual-route conditioning establishes explicit spatial correspondence without introducing an additional image-like mask input. We zero-initialize the final layers of \(P_{\mathrm{tar}}\) and \(P_{\mathrm{ctx}}\), so both residual branches start as no-ops and preserve the initial behavior of the pretrained model.


\subsubsection{Region-Balanced Flow Matching.}
The standard flow-matching objective uniformly averages velocity prediction errors over the target latent. In reference-guided image insertion, the insertion region is often small, allowing the much larger background to dominate optimization and
leaving the inserted subject’s appearance, structure, and placement under-supervised. A natural remedy is to assign a larger fixed weight to errors
within the insertion mask. Yet, when these errors are aggregated over the entire latent, their contribution still depends on mask size: given
comparable per-position errors, larger regions produce greater aggregate losses and gradients, while smaller regions remain underrepresented. Thus,
fixed foreground weighting cannot ensure consistent supervision across insertion objects of different scales.

To remove this scale dependency, we separately normalize the velocity
errors inside and outside the insertion region.
Let \(m_i\in[0,1]\) denote the insertion mask \(\mathbf{M}\) mapped to
the \(i\)-th target latent position.
We define the insertion-region and background losses as
\begin{align}
    \ell_{\mathrm{ins}}
    &=
    \frac{
        \sum_i
        m_i
        \left\|
            \mathbf{v}_{\theta,i}
            -
            \mathbf{u}_{t,i}
        \right\|_2^2
    }{
        \sum_i m_i
        +
        \varepsilon
    },
    \label{eq:insertion_region_fm}
    \\
    \ell_{\mathrm{bg}}
    &=
    \frac{
        \sum_i
        (1-m_i)
        \left\|
            \mathbf{v}_{\theta,i}
            -
            \mathbf{u}_{t,i}
        \right\|_2^2
    }{
        \sum_i (1-m_i)
        +
        \varepsilon
    },
    \label{eq:background_region_fm}
\end{align}
where the squared norm is computed over the latent channels at each
spatial position.
The resulting Region-Balanced Flow Matching objective is
\begin{equation}
    \mathcal{L}_{\mathrm{RBFM}}
    =
    \mathbb{E}_{\mathbf{z}_0,t,\boldsymbol{\epsilon}}
    \left[
        \frac{
            \lambda_{\mathrm{ins}}
            \ell_{\mathrm{ins}}
            +
            \lambda_{\mathrm{bg}}
            \ell_{\mathrm{bg}}
        }{
            \lambda_{\mathrm{ins}}
            +
            \lambda_{\mathrm{bg}}
        }
    \right],
    \label{eq:region_balanced_flow_matching}
\end{equation}
where \(\lambda_{\mathrm{ins}}\) and
\(\lambda_{\mathrm{bg}}\) control the relative importance of insertion
generation and background preservation.
Since each regional error is normalized by its own effective area, its
contribution no longer scales with the size of the corresponding region.
This provides stable supervision across insertion objects of different
scales while preventing the larger background from dominating the
optimization.

\subsubsection{Reference Classifier-Free Guidance.}
\label{sec:reference_cfg}

Reference-guided insertion requires the model to adapt the reference subject to the target scene while preserving its distinctive appearance and identity.
However, fine-grained reference cues may be weakened during this
adaptation, leading to texture degradation or identity drift.
We therefore introduce Reference Classifier-Free Guidance
(Reference CFG) to explicitly strengthen the influence of the visual reference.

During expert training, we randomly replace the reference-subject latent
\(\mathbf{z}^{\mathrm{ref}}\) and vision-language condition latent with a null condition with probability
\(p_{\mathrm{drop}}\), while keeping the geometry-conditioned target and source-context latents unchanged.
At inference, Reference CFG amplifies the velocity difference induced solely by the visual reference:
\begin{equation}
\begin{aligned}
    \mathbf{v}_{k}^{\mathrm{ref}}
    &=
    \mathbf{v}_{E_k}
    \left(
        \widetilde{\mathbf{z}}_t,
        t;
        \widetilde{\mathbf{z}}^{\mathrm{ctx}},
        \varnothing
    \right)
    \\
    &
    +
    s_r
    \Bigg[
        \mathbf{v}_{E_k}
        \left(
            \widetilde{\mathbf{z}}_t,
            t;
            \widetilde{\mathbf{z}}^{\mathrm{ctx}},
            \mathbf{z}^{\mathrm{ref}}
        \right)
        -
        \mathbf{v}_{E_k}
        \left(
            \widetilde{\mathbf{z}}_t,
            t;
            \widetilde{\mathbf{z}}^{\mathrm{ctx}},
            \varnothing
        \right)
    \Bigg],
\end{aligned}
\label{eq:reference_cfg_raw}
\end{equation}
where \(s_r\geq1\) controls the reference guidance strength.
Since the two predictions differ only in the reference condition, their difference isolates the reference-induced velocity direction.
Amplifying this direction improves the preservation of reference-specific appearance and identity cues.

\subsection{Insertion On-Policy Distillation}
\label{sec:iopd}

Insertion On-Policy Distillation (IOPD) consolidates the frozen
category-specific experts into a unified student \(S_{\theta}\),
initialized from the same pretrained model.
During distillation, category label \(y\) selects the matched expert
\(E_{k(y)}\).

The key idea of IOPD is to supervise the student on the states visited
along its own generation trajectory.
For compactness, we omit the fixed source-context and reference
conditions from the velocity notation.
Given an inference noise schedule
\(
\sigma_0>\sigma_1>\cdots>\sigma_N=0
\),
the student starts from Gaussian noise
\(
\mathbf{x}_0^{S}\sim\mathcal{N}(\mathbf{0},\mathbf{I})
\).
Let
\(
\widetilde{\mathbf{x}}_n^{S}
\)
denote the TAGC-augmented target input constructed from the current
student state
\(
\mathbf{x}_n^{S}
\).
The student then follows its own velocity field:
\begin{equation}
    \begin{gathered}
        \mathbf{x}_{n+1}^{S}
        =
        \mathbf{x}_n^{S}
        +
        \left(
            \sigma_{n+1}-\sigma_n
        \right)
        \mathbf{v}_{S_{\theta}}
        \left(
            \widetilde{\mathbf{x}}_n^{S},
            t_n
        \right),
        \\
        \text{for } n=0,\ldots,N-1.
    \end{gathered}
    \label{eq:iopd_rollout}
\end{equation}
The rollout is performed without gradient tracking and is used only to
collect the intermediate states visited by the current student.

For each student-visited state
\(
\mathbf{x}_n^{S}
\),
we query the matched expert
\(
E_{k(y)}
\)
at the same state and timestep.
The expert is evaluated with both the positive-reference and
null-reference conditions while keeping the source scene and TAGC
conditions unchanged.
These predictions are combined using Reference CFG in
Eqs.~\eqref{eq:reference_cfg_raw} to construct the reference-guided
expert target
\(
    \mathbf{v}_{E_{k(y)}}^{\mathrm{ref}}
    \left(
        \widetilde{\mathbf{x}}_n^{S},
        t_n
    \right).
\)
The student prediction is recomputed with gradients using only the
positive-reference condition.
The resulting IOPD objective is
\begin{equation}
\mathcal{L}_{\mathrm{IOPD}}
=
\mathbb{E}_{n}
\left[
\left\|
\mathbf{v}_{S_{\theta}}\!\left(\widetilde{\mathbf{x}}_{n}^{S}, t_{n}\right)
-
\operatorname{sg}\!\left[
\mathbf{v}_{E_{k(y)}}^{\mathrm{ref}}\!\left(\widetilde{\mathbf{x}}_{n}^{S}, t_{n}\right)
\right]
\right\|_{2}^{2}
\right],
\label{eq:iopd_objective}
\end{equation}
where
\(
\operatorname{sg}[\cdot]
\)
denotes stop-gradient, and the expectation is taken over training
examples and initial noise.

By matching the student to category-specific, reference-guided expert
velocities along its own generation trajectory, IOPD transfers both
specialized insertion behavior and strengthened reference fidelity into a single shared model.

\section{Experiments}

\begin{table*}[t]
    \centering

    \small
    \setlength{\tabcolsep}{4.0pt}
    \renewcommand{\arraystretch}{1.08}

    \begin{tabularx}{\textwidth}{
        @{}
        >{\raggedright\arraybackslash}p{0.27\textwidth}
        *{6}{>{\centering\arraybackslash}X}
        @{}
    }
        \toprule

        \textbf{Method}
        & \textbf{DINO-I}$\uparrow$
        & \textbf{CLIP-I}$\uparrow$
        & \textbf{PSNR}$\uparrow$
        & \textbf{SSIM}$\uparrow$
        & \textbf{LPIPS}$\downarrow$
        & \textbf{FID}$\downarrow$ \\

        \midrule
        \multicolumn{7}{c}{
            \textbf{\textit{(a) Insertion-Oriented Components}}
        } \\
        \cmidrule(lr){1-7}

        Baseline
        & 0.6889
        & 0.8926
        & 23.47
        & 0.8819
        & 0.0941
        & 1.58 \\

        w/ TAGC
        & 0.6972
        & 0.8949
        & 23.92
        & 0.8894
        & 0.0878
        & 1.51 \\

        w/ RBFM
        & 0.6961
        & 0.8944
        & 24.06
        & 0.8887
        & 0.0848
        & 1.46 \\

        w/ Reference CFG
        & 0.7076
        & 0.8999
        & 23.62
        & 0.8843
        & 0.0917
        & 1.55 \\

        \midrule
        \multicolumn{7}{c}{
            \textbf{\textit{(b) Multi-Category Training and Consolidation}}
        } \\
        \cmidrule(lr){1-7}

        Direct Joint Training
        & 0.6478
        & 0.8813
        & 19.36
        & 0.7894
        & 0.1746
        & 1.86 \\

        Category-Specific Experts
        & \textbf{0.7187}
        & \textbf{0.9031}
        & \underline{24.36}
        & \textbf{0.8991}
        & \textbf{0.0777}
        & \textbf{1.34} \\

        \textbf{InsertFuse (IOPD)}
        & \underline{0.7148}
        & \underline{0.9020}
        & \textbf{24.40}
        & \underline{0.8967}
        & \underline{0.0795}
        & \underline{1.39} \\

        \bottomrule
    \end{tabularx}

    \caption{
        \textbf{Ablation study on our multi-category test set.}
        In the first block, TAGC, Region-Balanced Flow Matching, and
        Reference CFG are independently added to the same baseline
        experts.
        In the second block, Direct Joint Training optimizes a single
        model on mixed-category data, Category-Specific Experts report
        the aggregated performance obtained using the matched expert for
        each category, and InsertFuse consolidates these experts into a
        unified student through IOPD.
    }
    \label{tab:overall_ablation}
\end{table*}


\begin{table}[t]
    \centering

    \small
    \renewcommand{\arraystretch}{1.12}
    \setlength{\tabcolsep}{3.0pt}

    \begin{tabularx}{\columnwidth}{
        @{}
        >{\raggedright\arraybackslash}p{0.34\columnwidth}
        *{3}{>{\centering\arraybackslash}X}
        @{}
    }
        \toprule

        \multirow{2}{*}{\textbf{Method}}
        & \multicolumn{3}{c}{\textbf{Preference Share (\%)}} \\

        \cmidrule(lr){2-4}

        &
        \makecell{\textbf{Reference}\\\textbf{Fidelity}}
        &
        \makecell{\textbf{Insertion}\\\textbf{Quality}}
        &
        \makecell{\textbf{Overall}\\\textbf{Quality}} \\

        \midrule

        AnyDoor
        & 1.6
        & 1.6
        & 1.6 \\

        Insert Anything
        & \underline{21.2}
        & \underline{17.0}
        & 14.0 \\

        A$^2$-Edit
        & 14.8
        & 13.2
        & \underline{18.4} \\

        \midrule

        FLUX.1 Kontext
        & 1.4
        & 1.8
        & 1.8 \\

        Qwen-Edit-2511
        & 5.2
        & 4.8
        & 5.0 \\

        Qwen-Image-2.0
        & 4.0
        & 3.4
        & 3.8 \\

        \midrule

        \textbf{InsertFuse (Ours)}
        & \textbf{51.8}
        & \textbf{58.2}
        & \textbf{55.4} \\

        \bottomrule
    \end{tabularx}

    \caption{
        Multi-way user-study results on our multi-category test set.
        For each example and evaluation criterion, participants select
        one preferred result from all anonymized methods.
        Each value denotes the share of valid selections received by the
        corresponding method.
    }
    \label{tab:user_study}
\end{table}

\begin{figure}[t]
    \centering
    \includegraphics[
        width=0.95\columnwidth
    ]{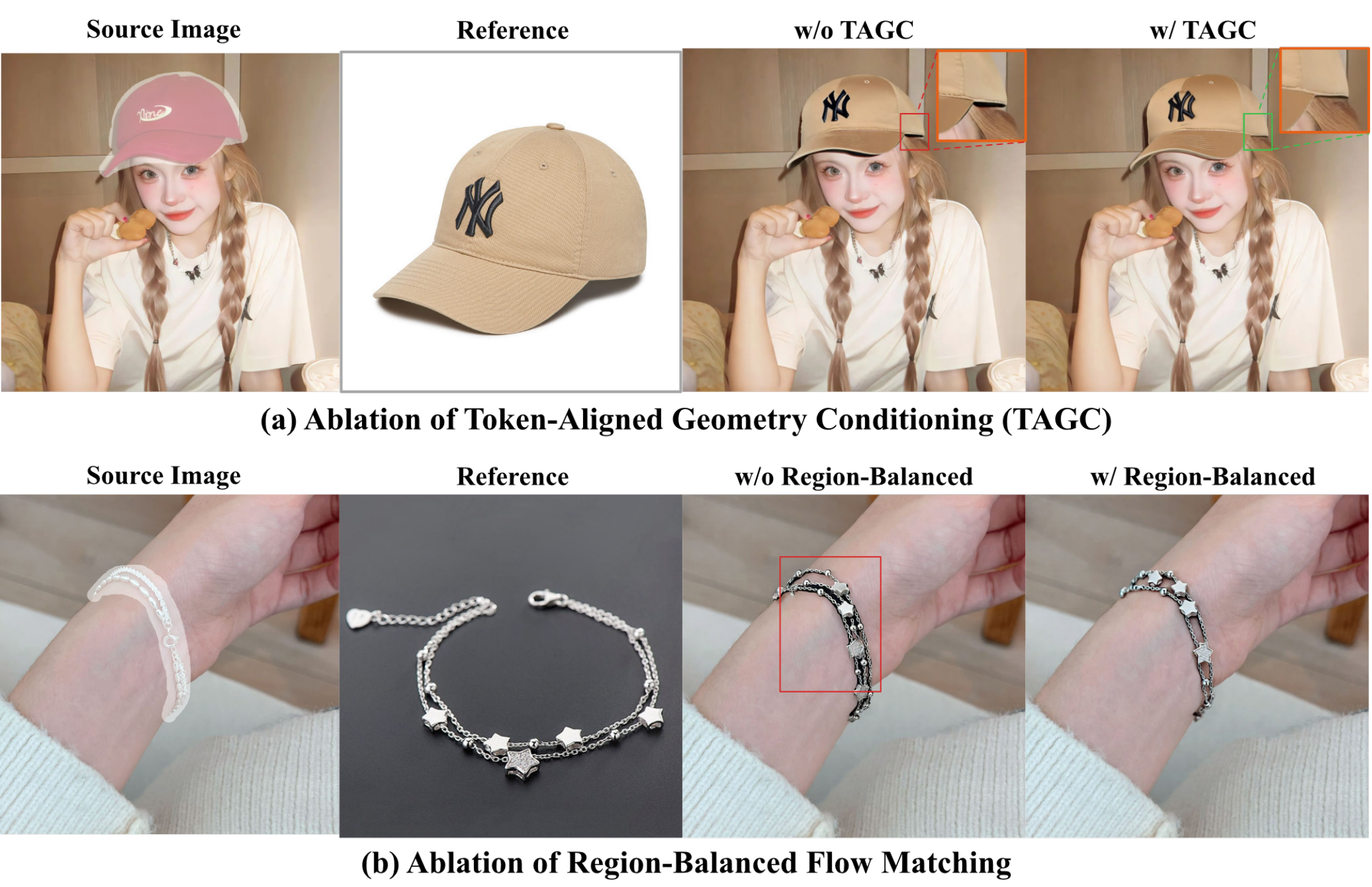}
    \caption{
        \textbf{Qualitative ablations of TAGC and Region-Balanced Flow
        Matching.}
        \textbf{(a)} TAGC improves geometric alignment and produces a cleaner transition between the cap and the head.
        \textbf{(b)} RBFM reduces bracelet artifacts and recovers clearer
        chain and star details.
    }
    \label{fig:tagc_rbfm_ablation}
\end{figure}

\begin{figure}[t]
    \centering
    \includegraphics[
        width=0.95\columnwidth
    ]{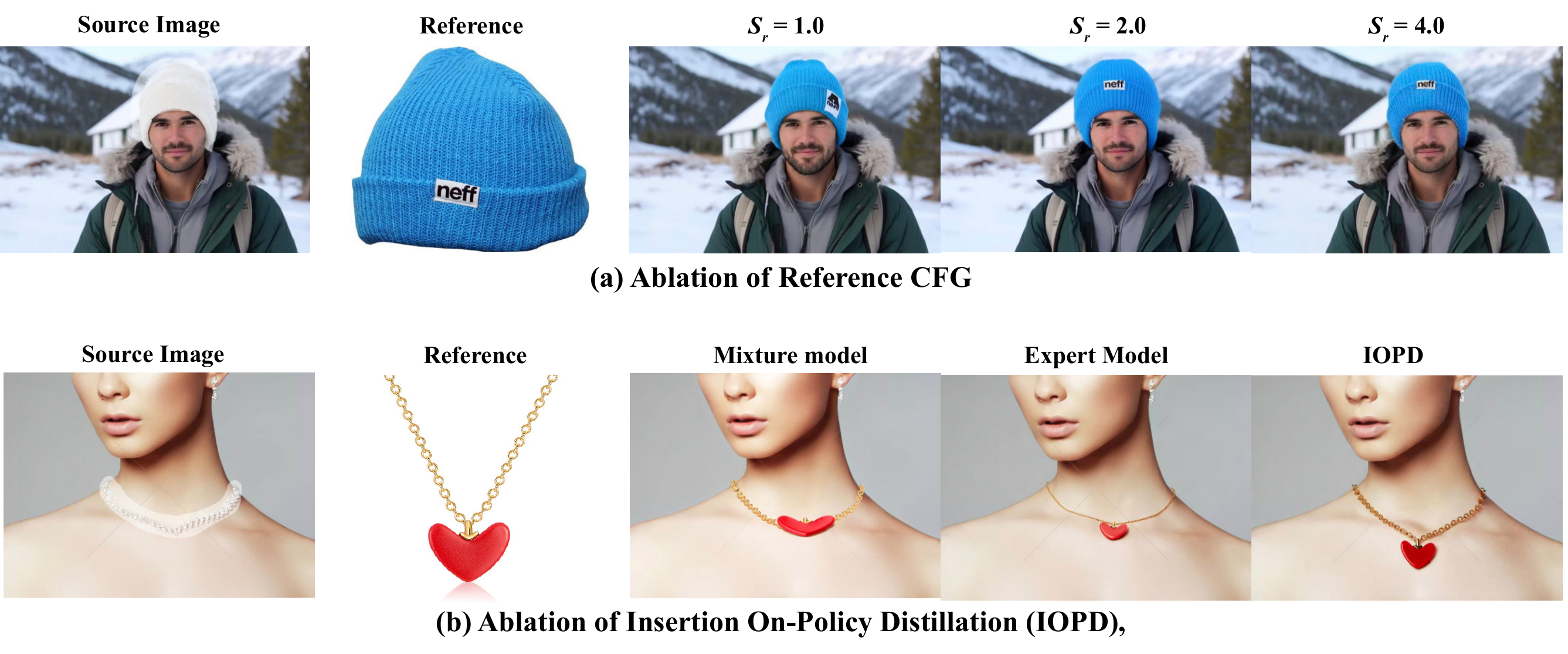}
    \caption{
        \textbf{Qualitative ablations of Reference CFG and IOPD.}
        \textbf{(a)} Compared with ordinary conditional prediction at
        \(s_r=1.0\), increasing the reference guidance strength improves
        the preservation of the beanie's texture and distinctive
        logo.
        \textbf{(b)} Direct joint training distorts the heart-shaped
        pendant, whereas the category-specific expert and IOPD produce
        a more reference-consistent necklace.
        IOPD transfers this specialized insertion behavior into a unified
        student.
    }
    \label{fig:cfg_iopd_ablation}
\end{figure}

\subsection{Experimental Setup}
\label{sec:experimental_setup}

\subsubsection{Implementation Details.}
We build InsertFuse upon Qwen-Image-Edit-2511~\cite{wu2025qwenimage} and keep the original
pretrained parameters frozen, while optimizing the LoRA adapters and the
lightweight geometry-conditioning modules in TAGC.
All images are processed under a pixel budget of approximately one
megapixel while preserving their original aspect ratios.
Both the rank and scaling factor of LoRA are set to 256.
In the first stage, we train five category-specific experts specializing
in accessory, animal, garment, general-object, and human insertion,
respectively.
Each expert is trained for five epochs.
The insertion and background weights in Region-Balanced Flow Matching
are set to
\(\lambda_{\mathrm{ins}}=3\) and
\(\lambda_{\mathrm{bg}}=1\), respectively, and the reference condition
is dropped with probability \(p_{\mathrm{drop}}=0.2\).
We then freeze the experts and initialize a unified student from the
same backbone.
The student performs 10-step deterministic on-policy rollouts, while
the matched expert constructs distillation targets using Reference CFG
with \(s_r=4.0\).
The student always uses a single positive-reference prediction.
All models are evaluated with 40 sampling steps, and all experiments
are conducted on eight NVIDIA H200 GPUs.
We provided detailed experimental settings in the supplement.


\subsubsection{Baselines and Datasets.}
For each category, we collect \(1{,}000\) training examples and construct
a dedicated test set. Their union forms our multi-category test set.
To further evaluate cross-dataset generalization, we additionally test
all applicable methods on the public AnyInsertion~\cite{song2025insertanything} benchmark.
We compare InsertFuse with representative reference-guided insertion
methods, including AnyDoor~\cite{chen2024anydoor},
Insert Anything~\cite{song2025insertanything}, and
A$^2$-Edit~\cite{zheng2026a2edit}, as well as recent unified image editing models, including FLUX.1 Kontext~\cite{labs2025flux1kontextflowmatching}, Qwen-Image-Edit-2511~\cite{wu2025qwenimage} and Qwen-Image-2.0~\cite{zhao2026qwenimage20technicalreport}.
We use official implementations with the recommended inference settings.

\subsubsection{Evaluation Metrics.}
Following prior reference-guided image insertion
studies~\cite{song2025insertanything,zheng2026a2edit}, we report
DINO-I~\cite{zhang2023magicbrush} and CLIP-I~\cite{gandikota2022imagereward} to evaluate reference fidelity;
PSNR, SSIM, and LPIPS~\cite{zhang2018unreasonable} to measure image-level reconstruction and
perceptual quality; and FID~\cite{heusel2017gans} to assess the distributional discrepancy
between generated and real target images.

\subsection{Analysis and Discussion}
\label{analysis_and_discussion}

\subsubsection{Quantitative Results.}
\label{sec:quantitative_results}

As shown in Tab.~\ref{tab:main_comparison}, InsertFuse achieves the best
results on most metrics across both evaluation sets and remains
second-best elsewhere.
Its DINO-I/CLIP-I scores of \(0.7838/0.9156\) on AnyInsertion and
\(0.7148/0.9020\) on our test set indicate strong reference fidelity.
PSNR/SSIM scores of \(24.26/0.8850\) and \(24.40/0.8967\), together
with low LPIPS and FID, further demonstrate accurate reconstruction and
strong perceptual quality across both external and category-diverse
settings.



\subsubsection{Qualitative Results.}
\label{sec:qualitative_results}

Fig.~\ref{fig:qualitative_comparison} compares different methods across
diverse insertion categories.
For garments, InsertFuse preserves the asymmetric graphic, central
panel, and waistband details while adapting the shorts naturally to the
body.
For the wardrobe, it not only reconstructs the characteristic paneled
appearance of the reference object, but also better preserves the
foreground lamp and the correct occlusion relationship.
For the watch, InsertFuse is the only method that correctly reorients
the object to match the wrist perspective while preserving its
distinctive dial and strap details.
It also maintains the reference identity and clothing appearance under
the target human pose, and accurately reconstructs the double-chain
layout and star-shaped ornaments in the small bracelet region.
In contrast, competing methods frequently distort reference-specific
details, produce structural artifacts, or fail to satisfy the target
geometry.

Overall, InsertFuse effectively combines reference preservation with
category-dependent adaptation to pose, perspective, occlusion, and
regional geometry.

\subsubsection{User Study.}
\label{sec:user_study}

We conduct a multi-way user study in which participants view the source
image, target mask, reference image, and randomly ordered anonymized
outputs, and select the best result for
\textit{reference fidelity} (appearance and identity),
\textit{insertion quality} (placement, adaptation, and blending), and
\textit{overall quality} (realism and preference).
Tab.~\ref{tab:user_study} reports the resulting selection shares.
InsertFuse receives the largest proportion of votes across the three
criteria, demonstrating stronger reference preservation, more natural
scene integration, and higher overall perceptual quality.


\subsection{Ablation Study}
\label{sec:ablation_study}

We conduct module-level ablations on our multi-category test set, with
the results reported in Tab.~\ref{tab:overall_ablation}.
For the first three ablations, we independently train five
category-specific experts and evaluate each category using its
corresponding teacher.
The baseline experts use the original conditioning formulation and
standard global flow matching.
Each component is then introduced independently while all other
training settings remain unchanged.

\subsubsection{Token-Aligned Geometry Conditioning.}
As shown in Tab.~\ref{tab:overall_ablation}, TAGC consistently improves
the quantitative performance over the baseline teachers.
In Fig.~\ref{fig:tagc_rbfm_ablation}(a), the model without TAGC produces
a visible discontinuity at the cap boundary, whereas TAGC yields better
geometric alignment and a cleaner transition with the surrounding
content.
These results demonstrate that token-aligned geometry conditioning
improves spatial control and boundary quality.

\subsubsection{Region-Balanced Flow Matching.}
As shown in Tab.~\ref{tab:overall_ablation}, RBFM provides consistent gains over standard global flow matching.
In Fig.~\ref{fig:tagc_rbfm_ablation}(b), global flow matching produces
entangled bracelet structures and unclear local details, while RBFM
generates a more coherent chain layout and clearer ornaments.
This confirms the effectiveness of balanced regional supervision,
particularly for small insertion regions.

\subsubsection{Reference CFG.}
As shown in Tab.~\ref{tab:overall_ablation}, Reference CFG consistently
improves reference-oriented metrics.
Fig.~\ref{fig:cfg_iopd_ablation}(a) further shows that stronger
reference guidance better preserves the knitted texture and distinctive
logo of the reference beanie.
These results validate Reference CFG for strengthening the reference
fidelity of category-specific experts.

\subsubsection{Insertion On-Policy Distillation.}
The results in
Tab.~\ref{tab:overall_ablation} and
Fig.~\ref{fig:cfg_iopd_ablation}(b) consistently show that direct joint
training suffers from cross-category interference, whereas IOPD
effectively consolidates the specialized capabilities of
category-specific experts into a unified student, leading to stronger
multi-category insertion performance.
\section{Conclusion}


We presented \textbf{InsertFuse}, a unified framework for
multi-category reference-guided image insertion.
By decoupling category-specific expert learning from cross-category
consolidation, InsertFuse employs Insertion On-Policy Distillation to
transfer specialized insertion policies and reference-enhanced expert
supervision into a single student, mitigating the interference caused by
direct joint training.
Token-Aligned Geometry Conditioning and Region-Balanced Flow Matching
provide explicit geometric guidance and balanced regional optimization,
while Reference CFG strengthens reference fidelity.
Extensive experiments across diverse categories and datasets demonstrate
strong performance in reference preservation, spatial adaptation, and
generation quality, achieving state-of-the-art results on most
evaluation metrics.
InsertFuse provides a unified and practical solution for diverse reference-guided image insertion tasks.


\bibliography{aaai2027}


\clearpage
\setcounter{page}{1}
\appendix

\section{Training and Implementation Details}
\label{supp:training_implementation}

\paragraph{Training overview.}
InsertFuse is trained in two stages.
In the first stage, we train five category-specific insertion experts
for accessory, animal, garment, general-object, and human insertion,
respectively.
Each expert is optimized only on its corresponding data subset, allowing
it to learn the appearance preservation, geometric adaptation, and
scene-composition patterns required by that category.
In the second stage, we freeze all category-specific experts and train a
unified student using Insertion On-Policy Distillation (IOPD).
The student generates its own deterministic trajectories, while the
expert associated with each training sample provides a Reference-CFG
distillation target at every visited trajectory state.
After distillation, only the unified student is retained for inference.

Text prompts are kept empty in the reported experiments so that insertion
is controlled exclusively by the source scene, reference appearance, and
spatial condition.

\paragraph{Image preprocessing.}
Target and reference images are processed under a pixel budget of
$1{,}048{,}576$ pixels while preserving their original aspect ratios.
The resulting spatial dimensions are adjusted to valid
model-compatible multiples required by the VAE and token-packing
operations.
RGB images are resized with bilinear interpolation, whereas binary masks
are resized with nearest-neighbor interpolation and thresholded again
after resizing.

For the reference input, we first compute the tight foreground bounding
box from $\mathbf{m}^{\mathrm{ref}}$ and expand it by a padding ratio of
$0.15$.
The reference crop has a minimum side length of $64$ pixels and is aligned
to a multiple of $16$ before being resized to the model input resolution.
This preprocessing reduces irrelevant reference background while retaining
a moderate amount of surrounding context.

\paragraph{Adaptive target cropping.}
Small insertion regions can occupy only a few target-image tokens when
the complete scene is directly resized to the one-megapixel budget.
We therefore perform adaptive target cropping before model
preprocessing.
Cropping is activated when the ground-truth insertion bounding box
occupies less than $10\%$ of the target image.
The crop size is determined according to the bounding-box area, with a
minimum crop-area ratio of $0.25$ and an additional context margin of
$0.20$.
The crop follows the target model aspect ratio and has a minimum side
length of $16$ pixels.
The same geometric transformation is applied to the target image,
ground-truth mask, and all target-aligned conditions.
During inference, the generated crop is mapped back to the original
coordinate system and composited with the original image using the
ground-truth insertion mask.

\paragraph{Dual-mask construction and augmentation.}
We distinguish the ground-truth insertion mask
$\mathbf{m}^{\mathrm{gt}}$ from the condition mask
$\mathbf{m}^{\mathrm{cond}}$.
The former defines the region used by the training losses and final
compositing, whereas the latter is used only to construct the masked
source image and the TAGC geometry condition:
\begin{equation}
    \mathbf{x}^{\mathrm{src}}
    =
    \mathbf{x}^{\mathrm{tar}}
    \odot
    \left(
        1-\mathbf{m}^{\mathrm{cond}}
    \right).
\end{equation}
Consequently, inaccuracies or additional context in the condition mask do
not alter the supervision region.

During training, $\mathbf{m}^{\mathrm{cond}}$ is sampled using the
MaskMix augmentation strategy.
We use the exact ground-truth mask with probability $0.60$, a random
morphological mask with probability $0.20$, an augmented bounding-box mask
with probability $0.10$, and an asymmetric bounding-box mask with
probability $0.10$.
The morphological radius is sampled between $2\%$ and $12\%$ of the
object scale and clipped to $[1,32]$ pixels, with an anisotropy factor
between $1.5$ and $3.0$.
Bounding-box perturbations use an object-relative magnitude between
$0.05$ and $0.20$, clipped to $[2,64]$ pixels, while the asymmetric
box margins are sampled from $[0,0.10]$.
All augmented masks are constrained to cover
$\mathbf{m}^{\mathrm{gt}}$.
When the resulting condition mask covers more than $80\%$ of the target
image, we fall back to the ground-truth mask to prevent excessive removal
of source-scene context.
A single sampled condition mask is shared by the student and all expert
branches for the corresponding training sample.

\paragraph{TAGC implementation.}
TAGC converts the binary condition mask into token-aligned geometric
features.
Given $\mathbf{m}^{\mathrm{cond}}$, we compute a signed Euclidean distance
map $\mathbf{D}$, whose values are positive inside the masked region and
negative outside.
The input geometry representation is
\begin{equation}
    \mathbf{G}
    =
    \left[
        \mathbf{m}^{\mathrm{cond}},
        \tanh\left(
            \frac{\mathbf{D}}{\sigma_g}
        \right),
        \exp\left(
            -\frac{|\mathbf{D}|}{\sigma_g}
        \right)
    \right],
    \qquad
    \sigma_g=16.
\end{equation}
The three channels respectively encode region occupancy, signed
inside--outside distance, and boundary proximity.

The geometry tokenizer consists of a $3\times3$ convolution that projects
the three-channel input to $32$ hidden channels, a SiLU activation, and a
$1\times1$ output projection whose output dimension matches the target
image-token dimension.
The output projection is initialized to zero, making TAGC an exact
no-op at initialization and preserving the behavior of the pretrained
backbone.
The geometry features are aligned with the packed target-image token grid
and added residually to the target token stream and context token stream.
A separate geometry tokenizer is learned for each category expert in
Stage I.
During IOPD, the expert geometry tokenizers are frozen, while the unified
student learns its own geometry tokenizer.

\paragraph{Trainable parameterization.}
We initialize all models from Qwen-Image-Edit-2511.
The pretrained VAE, visual-language condition encoder, and Transformer
backbone remain frozen throughout training.
LoRA adapters are inserted into the query, key, value, and output
projections of the joint-attention layers, the additional image-condition
attention projections, and the input and output projections of the image
and text MLPs.
Both the LoRA rank and LoRA scaling factor are set to $256$, corresponding
to an effective multiplier of $\alpha/r=1$.
The LoRA parameters are initialized with a Gaussian initialization.
Thus, the only trainable components in Stage I are the expert-specific
LoRA adapters and TAGC geometry tokenizer.
In Stage II, only the unified student's LoRA adapters and geometry
tokenizer are optimized; all five experts and their associated geometry
tokenizers are kept frozen.

\begin{table*}[t]
    \centering
    \small
    \label{tab:supp_training_hyperparams}
    \setlength{\tabcolsep}{5.0pt}
    \renewcommand{\arraystretch}{1.12}
    \begin{tabular}{
        p{0.27\textwidth}
        p{0.32\textwidth}
        p{0.32\textwidth}
    }
        \toprule
        \textbf{Configuration}
        & \textbf{Stage I: Category-Specific Experts}
        & \textbf{Stage II: Unified IOPD} \\
        \midrule
        Backbone
        & Qwen-Image-Edit-2511
        & Qwen-Image-Edit-2511 \\

        Number of models
        & Five independently trained experts
        & One unified student with five frozen experts \\

        Trainable parameters
        & Expert LoRA and expert TAGC
        & Student LoRA and student TAGC \\

        LoRA rank / scaling factor
        & $256/256$
        & $256/256$ \\

        Main objective
        & Region-Balanced Flow Matching and region reconstruction
        & IOPD velocity matching \\


        Optimizer
        & AdamW
        & AdamW \\

        Learning rate
        & $1\times10^{-4}$
        & $1\times10^{-4}$ \\

        AdamW $(\beta_1,\beta_2)$
        & $(0.9,0.999)$
        & $(0.9,0.999)$ \\

        Weight decay
        & $0.01$
        & $0.01$ \\

        Learning-rate schedule
        & Cosine
        & Cosine \\

        Warm-up iterations
        & $500$
        & $500$ \\

        Per-GPU batch size
        & $1$
        & $1$ \\

        Gradient accumulation
        & $1$
        & $5$ \\

        Effective global batch size
        & $8$
        & $40$ \\

        Maximum gradient norm
        & $1.0$
        & $1.0$ \\

        Reference handling
        & Reference dropout with $p_{\mathrm{drop}}=0.2$
        & Expert Reference CFG with $s_r=4.0$;
          one positive student prediction \\

        Insertion/background weights
        & $\lambda_{\mathrm{ins}}=3$,
          $\lambda_{\mathrm{bg}}=1$
        & Not used \\

        Region reconstruction weights
        & $\lambda_{\mathrm{LPIPS}}=0.5$,
          $\lambda_{\mathrm{bg\text{-}L1}}=0.5$
        & Not used \\

        On-policy rollout steps
        & Not applicable
        & $10$ \\

        Precision
        & BF16
        & BF16 \\

        GPUs
        & $8\times$ NVIDIA H200
        & $8\times$ NVIDIA H200 \\

        Evaluation steps
        & $40$
        & $40$ \\
        \bottomrule
    \end{tabular}
        \caption{
        Training hyperparameters of InsertFuse.
        Stage I trains the five category-specific experts independently,
        whereas Stage II distills their insertion capabilities into one
        unified student.
    }
\end{table*}

\paragraph{Optimization details.}
Both training stages use AdamW with
$\beta_1=0.9$, $\beta_2=0.999$, weight decay $0.01$, and
$\epsilon=10^{-8}$.
The learning rate is set to $1\times10^{-4}$ and follows a cosine schedule
with $500$ warm-up iterations.
We use BF16 mixed-precision training, gradient checkpointing, and a
maximum gradient norm of $1.0$.
The per-GPU batch size is $1$ in both stages.
Stage I uses no gradient accumulation, giving a global batch size of $8$.
Stage II accumulates gradients over $5$ micro-batches, giving an effective
global batch size of $40$.
All experiments are conducted on eight NVIDIA H200 GPUs.

\begin{table*}[t]
    \centering

    \small
    \setlength{\tabcolsep}{4.0pt}
    \renewcommand{\arraystretch}{1.08}

    \begin{tabularx}{\textwidth}{
        @{}
        >{\raggedright\arraybackslash}p{0.27\textwidth}
        *{6}{>{\centering\arraybackslash}X}
        @{}
    }
        \toprule

        \textbf{Method}
        & \textbf{DINO-I}$\uparrow$
        & \textbf{CLIP-I}$\uparrow$
        & \textbf{PSNR}$\uparrow$
        & \textbf{SSIM}$\uparrow$
        & \textbf{LPIPS}$\downarrow$
        & \textbf{FID}$\downarrow$ \\
        \midrule

        Baseline
        & 0.6889
        & 0.8926
        & 23.47
        & 0.8819
        & 0.0941
        & 1.58 \\

        Mask Latent Concatenation$^{\dagger}$
        & 0.6924
        & 0.8938
        & 23.71
        & 0.8846
        & 0.0916
        & \underline{1.54} \\

        Raw Mask Injection$^{\dagger}$
        & \underline{0.6956}
        & \underline{0.8943}
        & \underline{23.81}
        & \underline{0.8875}
        & \underline{0.0891}
        & 1.55 \\

        \midrule

        \textbf{TAGC (Ours)}
        & \textbf{0.6972}
        & \textbf{0.8949}
        & \textbf{23.92}
        & \textbf{0.8894}
        & \textbf{0.0878}
        & \textbf{1.51} \\

        \bottomrule
    \end{tabularx}

    \caption{
        \textbf{Comparison of different mask-conditioning strategies.}
        All variants use the same category-specific expert training
        setting and differ only in the representation and injection of
        the insertion mask.
        Mask Latent Concatenation encodes the binary mask through the
        frozen VAE and concatenates the resulting latent tokens with the
        visual conditions.
        Raw Mask Injection directly maps the binary mask to the visual
        token grid, while TAGC additionally incorporates signed-distance
        and boundary-proximity cues.
        Results are averaged over the five category-specific test sets.
        Entries marked with \(^{\dagger}\) are provisional estimates for
        experiment planning and will be replaced with measured results.
    }
    \label{tab:supp_tagc_conditioning}
\end{table*}

\section{Detailed Analysis of Token-Aligned Geometry Conditioning}
\label{supp:tagc_analysis}

\paragraph{Alternative mask-conditioning strategies.}
To determine whether the improvement of TAGC arises merely from
providing an explicit insertion mask or from its token-aligned geometric
representation, we compare four mask-conditioning strategies.
All variants are trained using the same category-specific expert
setting, training data, pretrained backbone, LoRA configuration,
optimization schedule, and training objective.
Only the representation and injection of the insertion mask are changed.

The \textit{Baseline} does not introduce an additional mask-conditioning
branch.
It receives the masked source-context image
$\mathbf{I}^{\mathrm{ctx}}$ and must infer the insertion region implicitly
from the missing content in the source scene.

For \textit{Mask Latent Concatenation}, we repeat the binary insertion
mask along the channel dimension to construct an RGB mask image and
encode it using the frozen VAE:
\begin{equation}
    \mathbf{z}^{\mathrm{mask}}
    =
    \mathcal{V}_{\mathrm{enc}}
    \left(
        \operatorname{Repeat}_{3}
        \left(
            \mathbf{M}
        \right)
    \right).
    \label{eq:supp_mask_latent}
\end{equation}
The resulting mask latent is packed in the same manner as the other
visual controls and concatenated with the visual-conditioning token
sequence.
This variant provides an explicit mask condition through the existing
VAE pathway, but the binary spatial structure is encoded using an
appearance-oriented image encoder, and its correspondence with the
target tokens must still be recovered by the generative backbone.

For \textit{Raw Mask Injection}, we remove the VAE encoding and directly
use the original binary mask as the spatial condition.
Specifically, the mask is resized to the model token grid and mapped to
the hidden dimension using a lightweight mask tokenizer:
\begin{equation}
    \mathbf{e}^{\mathrm{mask}}
    =
    \Phi_{\mathrm{mask}}
    \left(
        \mathbf{M}
    \right).
    \label{eq:supp_raw_mask_feature}
\end{equation}
We then apply the same branch-specific projections and residual injection
positions as TAGC:
\begin{equation}
\begin{aligned}
    \widetilde{\mathbf{z}}_{t}
    &=
    \mathbf{z}_{t}
    +
    P_{\mathrm{tar}}
    \left(
        \mathbf{e}^{\mathrm{mask}}
    \right),
    \\
    \widetilde{\mathbf{z}}^{\mathrm{ctx}}
    &=
    \mathbf{z}^{\mathrm{ctx}}
    +
    P_{\mathrm{ctx}}
    \left(
        \mathbf{e}^{\mathrm{mask}}
    \right).
\end{aligned}
\label{eq:supp_raw_mask_injection}
\end{equation}
The output projections are zero-initialized following TAGC.
This variant establishes direct spatial correspondence between the
binary mask and the visual tokens, but communicates only region
occupancy and contains no explicit distance or boundary information.

Finally, \textit{TAGC} uses the complete geometric representation
introduced in the main paper:
\begin{equation}
    \mathbf{G}
    \left(
        \mathbf{M}
    \right)
    =
    \left[
        \mathbf{M},
        \tanh
        \left(
            \frac{
                D_{\mathbf{M}}
            }{\tau}
        \right),
        \exp
        \left(
            -
            \frac{
                \left|
                    D_{\mathbf{M}}
                \right|
            }{\tau}
        \right)
    \right].
    \label{eq:supp_full_tagc_geometry}
\end{equation}
Compared with Raw Mask Injection, TAGC additionally exposes the signed
interior--exterior distance and boundary proximity of every spatial
location, while retaining the same token-aligned injection mechanism.

\paragraph{Results.}
As shown in Tab.~\ref{tab:supp_tagc_conditioning}, explicitly providing
the insertion mask offers more direct spatial supervision than requiring
the baseline to infer the editable region solely from the masked source
context.
However, Mask Latent Concatenation remains limited because the binary
mask is processed by the image VAE and introduced as a separate visual
condition, leaving the correspondence between mask and target tokens to
be learned implicitly.
Directly injecting the raw mask establishes explicit token-level
correspondence and therefore provides more effective spatial control,
but it only distinguishes the inside and outside of the insertion
region.

TAGC achieves the strongest overall performance by combining direct
token alignment with richer geometric cues.
The signed-distance channel informs each token of its relative position
with respect to the insertion boundary, while the boundary-proximity
channel explicitly identifies locations requiring accurate transitions
between generated and preserved content.
These results show that the gains of TAGC do not arise merely from adding
an explicit mask input; they primarily result from representing
boundary-aware mask geometry at the corresponding visual-token
locations.

\section{Detailed Analysis of Region-Balanced Flow Matching}
\label{supp:rbfm_analysis}

\paragraph{Comparison with fixed regional weighting.}
To further analyze the effectiveness of Region-Balanced Flow Matching
(RBFM), we compare it with standard global flow matching and a
straightforward fixed-weighting strategy.
All variants use the same category-specific expert training setting and
differ only in how the token-wise flow-matching errors are aggregated.
TAGC and Reference CFG are not included in this comparison.

Let
\begin{equation}
    e_i
    =
    \left\|
        \mathbf{v}_{\theta,i}
        -
        \mathbf{u}_{t,i}
    \right\|_2^2
\end{equation}
denote the velocity prediction error at the \(i\)-th target latent
position.
The standard global flow-matching objective uniformly averages the
errors over all target positions:
\begin{equation}
    \mathcal{L}_{\mathrm{Global}}
    =
    \mathbb{E}
    \left[
        \frac{1}{N}
        \sum_{i=1}^{N}
        e_i
    \right].
    \label{eq:supp_global_fm}
\end{equation}

A straightforward alternative is to assign a larger fixed weight to the
positions inside the insertion region.
We implement this strategy as
\begin{equation}
    \mathcal{L}_{\mathrm{Fixed}}
    =
    \mathbb{E}
    \left[
        \frac{
            \sum_i
            \left[
                \lambda_{\mathrm{ins}}m_i
                +
                \lambda_{\mathrm{bg}}(1-m_i)
            \right]
            e_i
        }{
            \sum_i
            \left[
                \lambda_{\mathrm{ins}}m_i
                +
                \lambda_{\mathrm{bg}}(1-m_i)
            \right]
            +
            \varepsilon
        }
    \right],
    \label{eq:supp_fixed_weighting}
\end{equation}
where \(m_i\in[0,1]\) denotes the insertion-mask value at the
corresponding latent position.
We use the same weights as RBFM,
\(\lambda_{\mathrm{ins}}=3\) and
\(\lambda_{\mathrm{bg}}=1\), to ensure a controlled comparison.

Although Fixed Weighting increases the per-position importance of the
insertion region, it still aggregates foreground and background errors
jointly.
Let
\(
\rho=\frac{1}{N}\sum_i m_i
\)
denote the insertion-area ratio.
The effective coefficient assigned to the insertion region under
Fixed Weighting is
\begin{equation}
    r_{\mathrm{ins}}^{\mathrm{Fixed}}
    =
    \frac{
        \lambda_{\mathrm{ins}}\rho
    }{
        \lambda_{\mathrm{ins}}\rho
        +
        \lambda_{\mathrm{bg}}(1-\rho)
    }.
    \label{eq:supp_fixed_effective_ratio}
\end{equation}
This coefficient decreases with the insertion-area ratio and remains
small when the insertion region occupies only a small portion of the
image.
Therefore, fixed foreground weighting does not fundamentally remove the
scale dependency of global flow matching.

In contrast, RBFM independently normalizes the insertion and background
errors by their corresponding effective areas before combining them:
\begin{equation}
    \mathcal{L}_{\mathrm{RBFM}}
    =
    \mathbb{E}
    \left[
        \frac{
            \lambda_{\mathrm{ins}}
            \ell_{\mathrm{ins}}
            +
            \lambda_{\mathrm{bg}}
            \ell_{\mathrm{bg}}
        }{
            \lambda_{\mathrm{ins}}
            +
            \lambda_{\mathrm{bg}}
        }
    \right].
\end{equation}
Consequently, the explicit regional coefficients are independent of the
mask area, allowing small insertion regions to receive stable
supervision without being overwhelmed by the substantially larger
background.

\begin{table*}[t]
    \centering

    \small
    \setlength{\tabcolsep}{4.0pt}
    \renewcommand{\arraystretch}{1.08}

    \begin{tabularx}{\textwidth}{
        @{}
        >{\raggedright\arraybackslash}p{0.27\textwidth}
        *{6}{>{\centering\arraybackslash}X}
        @{}
    }
        \toprule

        \textbf{Method}
        & \textbf{DINO-I}$\uparrow$
        & \textbf{CLIP-I}$\uparrow$
        & \textbf{PSNR}$\uparrow$
        & \textbf{SSIM}$\uparrow$
        & \textbf{LPIPS}$\downarrow$
        & \textbf{FID}$\downarrow$ \\
        \midrule

        Global FM (Baseline)
        & 0.6889
        & 0.8926
        & 23.47
        & 0.8819
        & 0.0941
        & \underline{1.58} \\

        Fixed Weighting$^{\dagger}$
        & \underline{0.6915}
        & \underline{0.8930}
        & \underline{23.51}
        & \underline{0.8826}
        & \underline{0.0934}
        & 1.59 \\

        \midrule

        \textbf{RBFM (Ours)}
        & \textbf{0.6961}
        & \textbf{0.8944}
        & \textbf{24.06}
        & \textbf{0.8887}
        & \textbf{0.0848}
        & \textbf{1.46} \\

        \bottomrule
    \end{tabularx}

    \caption{
        \textbf{Comparison of different flow-matching weighting
        strategies.}
        All variants use the same category-specific expert training
        setting and differ only in the aggregation of token-wise velocity
        errors.
        Global FM uniformly averages errors over the complete target
        latent.
        Fixed Weighting assigns larger per-position weights to the
        insertion region but retains a single global normalization.
        RBFM independently normalizes the insertion and background errors
        before combining them.
        Both Fixed Weighting and RBFM use
        \(\lambda_{\mathrm{ins}}=3\) and
        \(\lambda_{\mathrm{bg}}=1\).
        Results are averaged over the five category-specific test sets.
        The entry marked with \(^{\dagger}\) is a provisional estimate for
        experiment planning and must be replaced with the measured result.
    }
    \label{tab:supp_rbfm_weighting}
\end{table*}

\paragraph{Results.}
As shown in Tab.~\ref{tab:supp_rbfm_weighting}, Fixed Weighting provides
only marginal changes over standard global flow matching.
Although assigning a larger foreground weight slightly improves
reference fidelity and local perceptual reconstruction, the overall
PSNR and SSIM remain close to the baseline, while FID exhibits a small
fluctuation.
This is because the total contribution of the insertion region remains
coupled to its spatial area under a single global normalization.

RBFM yields substantially stronger improvements across all evaluation
metrics.
In particular, it improves DINO-I and CLIP-I while increasing PSNR and
SSIM and noticeably reducing LPIPS and FID.
These results demonstrate that the effectiveness of RBFM does not arise
merely from assigning a larger foreground weight.
Instead, independently normalizing the insertion and background errors
is essential for providing scale-independent supervision and preventing
small insertion regions from being dominated by the background.

\section{Additional Analysis of Insertion On-Policy Distillation}
\label{supp:iopd_analysis}

\subsection{Cross-Category Gradient Interference in Direct Joint Training}
\label{supp:cross_category_gradient}

\paragraph{Motivation.}
The main paper shows that directly training a single model on mixed-category
insertion data substantially underperforms both the category-specific
experts and the unified student obtained through IOPD.
To examine the optimization behavior underlying this degradation, we
analyze the gradient compatibility among the five insertion categories
used in our experiments: accessory, garment, human, animal, and
general-object insertion.

In contrast to evaluating only the final performance of direct joint
training, this analysis measures whether different category objectives
provide compatible update directions in the shared parameter space.
If two category gradients have a negative inner product, an update that
reduces the loss of one category increases the loss of the other under a
first-order approximation.
This provides a direct optimization-level indicator of cross-category
interference.

\paragraph{Analysis protocol.}
We conduct the analysis at a fixed checkpoint of the directly jointly
trained model.
For each repeated measurement, we independently construct a batch for
each insertion category and compute its category-mean gradient with
respect to the shared LoRA parameters:
\begin{equation}
    \mathbf{g}_{c}^{(r)}
    =
    \nabla_{\theta_{\mathrm{LoRA}}}
    \mathcal{L}_{c}^{(r)},
    \label{eq:supp_category_gradient}
\end{equation}
where \(c\) denotes the insertion category and \(r\) indexes repeated
measurements.
All category gradients within the same repeat are evaluated at the same
model parameters and use the same training objective.

For each category pair \((i,j)\), we compute the cosine similarity
\begin{equation}
    s_{ij}^{(r)}
    =
    \frac{
        \left(
            \mathbf{g}_{i}^{(r)}
        \right)^{\top}
        \mathbf{g}_{j}^{(r)}
    }{
        \left\|
            \mathbf{g}_{i}^{(r)}
        \right\|_2
        \left\|
            \mathbf{g}_{j}^{(r)}
        \right\|_2
        +
        \varepsilon
    }.
    \label{eq:supp_gradient_cosine}
\end{equation}
The reported mean cosine similarity is
\begin{equation}
    \overline{s}_{ij}
    =
    \frac{1}{R}
    \sum_{r=1}^{R}
    s_{ij}^{(r)},
    \label{eq:supp_mean_gradient_cosine}
\end{equation}
where \(R\) is the number of repeated measurements.
We additionally report the fraction of measurements exhibiting negative
gradient alignment:
\begin{equation}
    q_{ij}
    =
    \frac{1}{R}
    \sum_{r=1}^{R}
    \mathbf{1}
    \left[
        s_{ij}^{(r)}<0
    \right].
    \label{eq:supp_negative_fraction}
\end{equation}
Finally, we measure the category-wise gradient magnitude using
\begin{equation}
    a_c^{(r)}
    =
    \left\|
        \mathbf{g}_{c}^{(r)}
    \right\|_2.
    \label{eq:supp_gradient_magnitude}
\end{equation}

The interpretation of negative gradient alignment follows directly from
a first-order Taylor expansion.
After applying an update along the gradient of category \(i\), the change
in the loss of category \(j\) is approximately
\begin{equation}
\begin{aligned}
    &
    \mathcal{L}_{j}
    \left(
        \theta
        -
        \eta\mathbf{g}_{i}
    \right)
    -
    \mathcal{L}_{j}(\theta)
    \\
    &\qquad\approx
    -
    \eta
    \mathbf{g}_{j}^{\top}
    \mathbf{g}_{i}.
\end{aligned}
\label{eq:supp_first_order_interference}
\end{equation}
Therefore, when
\(
\mathbf{g}_{j}^{\top}\mathbf{g}_{i}<0
\),
the update for category \(i\) increases the objective of category \(j\)
to first order.

\begin{figure*}[t]
    \centering

    \begin{minipage}[t]{0.315\textwidth}
        \centering
        \includegraphics[
            width=\linewidth
        ]{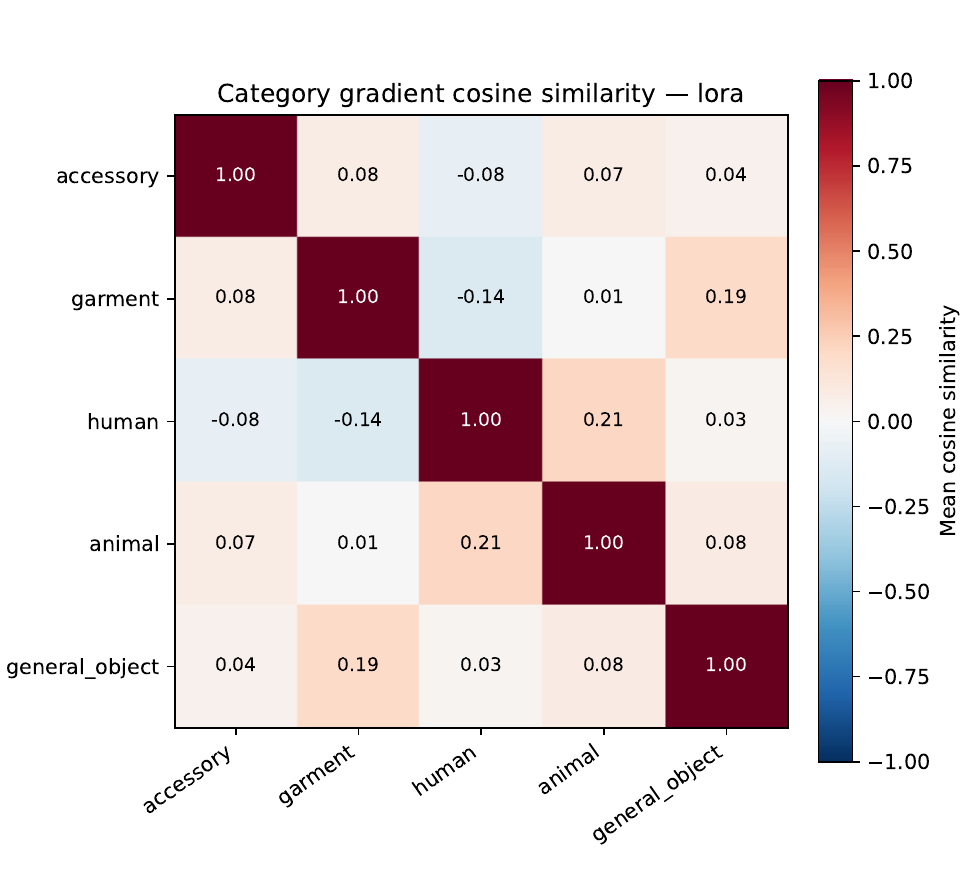}
        \vspace{-2mm}

        {\small\textbf{(a)} Mean gradient cosine similarity}
    \end{minipage}
    \hfill
    \begin{minipage}[t]{0.315\textwidth}
        \centering
        \includegraphics[
            width=\linewidth
        ]{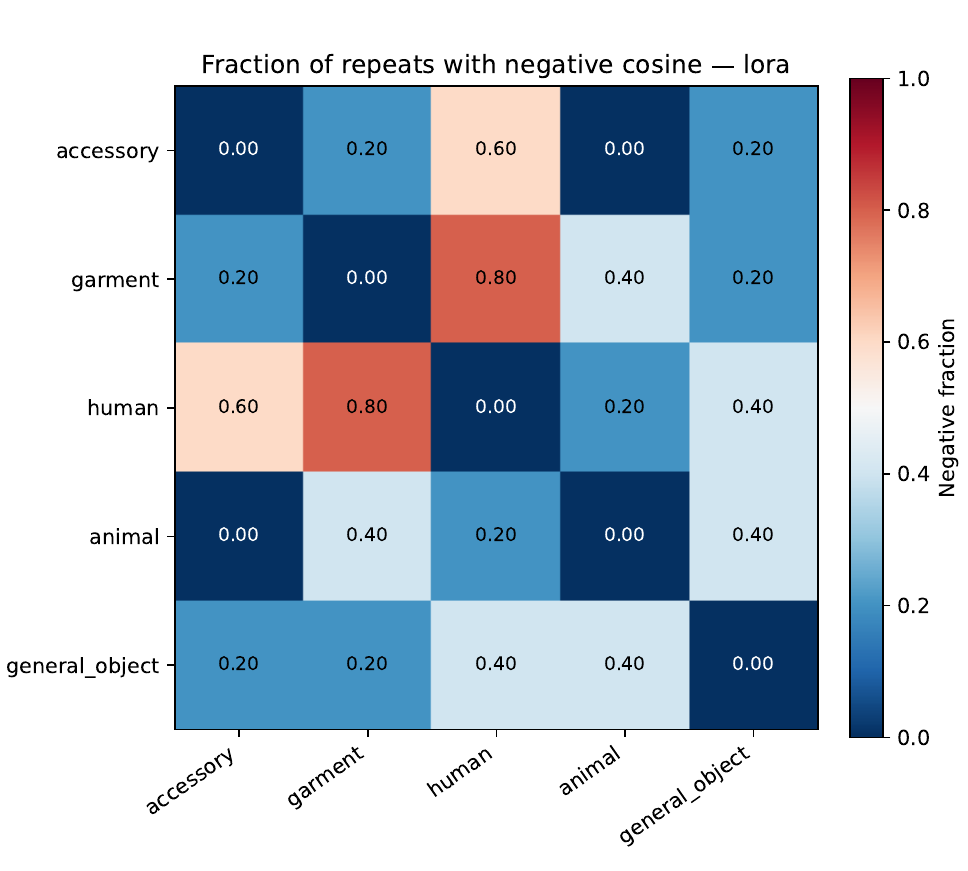}
        \vspace{-2mm}

        {\small\textbf{(b)} Fraction of negative measurements}
    \end{minipage}
    \hfill
    \begin{minipage}[t]{0.350\textwidth}
        \centering
        \includegraphics[
            width=\linewidth
        ]{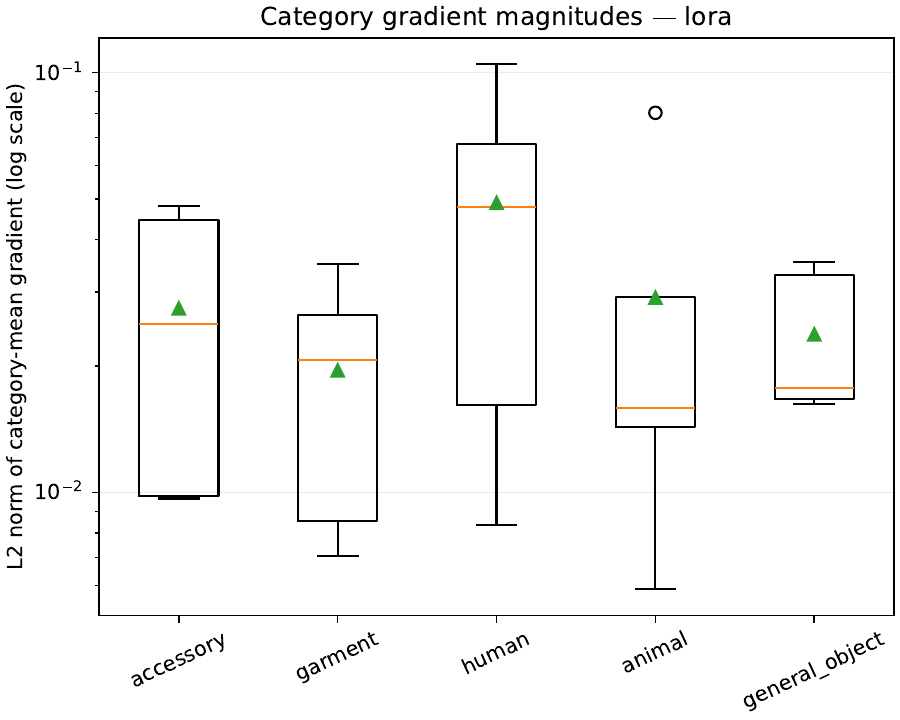}
        \vspace{-2mm}

        {\small\textbf{(c)} Category-gradient magnitudes}
    \end{minipage}

    \caption{
        \textbf{Cross-category gradient analysis under direct joint
        training.}
        \textbf{(a)} Mean pairwise cosine similarity between category
        gradients computed over the shared LoRA parameters.
        Values close to zero indicate weakly aligned update directions,
        while negative values indicate first-order gradient interference.
        \textbf{(b)} Fraction of repeated measurements for which each
        category pair exhibits negative cosine similarity.
        Human--garment and human--accessory show the most recurrent
        conflicts.
        \textbf{(c)} Distribution of the
        \(\ell_2\) norms of category-mean gradients, displayed on a
        logarithmic scale; green triangles denote the means.
        The results reveal both heterogeneous gradient directions and
        unequal update magnitudes across insertion categories.
    }
    \label{fig:supp_cross_category_gradient}
\end{figure*}

\paragraph{Weak and heterogeneous gradient alignment.}
As shown in
Fig.~\ref{fig:supp_cross_category_gradient}(a), the mean off-diagonal
cosine similarity across the ten category pairs is approximately
\(0.05\).
Moreover, seven of the ten pairs have an absolute cosine similarity
below \(0.1\).
Thus, most categories do not provide strongly aligned update directions,
even when their losses are evaluated at the same model parameters.

The compatibility between categories is also highly heterogeneous.
Human--garment and human--accessory exhibit negative mean cosine
similarities of \(-0.14\) and \(-0.08\), respectively, indicating that
their average update directions interfere with each other.
In contrast, human--animal and garment--general-object obtain positive
cosine similarities of \(0.21\) and \(0.19\), respectively.
These results show that the issue is not universal antagonism among all
categories.
Rather, different insertion categories exhibit substantially different
levels of optimization compatibility, making a single shared update
direction difficult to maintain consistently.

\paragraph{Recurrent pairwise conflicts.}
A negative average similarity alone could be caused by a small number of
noisy batches.
We therefore further examine how frequently each category pair exhibits
negative alignment across repeated measurements.
As shown in
Fig.~\ref{fig:supp_cross_category_gradient}(b), human--garment produces
negative cosine similarity in \(80\%\) of the measurements, while
human--accessory is negative in \(60\%\) of the measurements.
Garment--animal, human--general-object, and
animal--general-object also exhibit negative alignment in \(40\%\) of
the measurements.

Averaged over the ten non-repeated category pairs, approximately
\(34\%\) of the measured pairwise relationships are negative.
Furthermore, nine of the ten category pairs exhibit negative alignment
in at least one repeated measurement.
The negative relationships are therefore not restricted to a single
outlier batch.
Instead, direct joint training repeatedly encounters incompatible update
directions for several category combinations, with human--garment and
human--accessory showing the clearest conflicts.

\paragraph{Unequal gradient magnitudes.}
Besides directional compatibility, the contribution of each category to
the joint update also depends on its gradient magnitude.
As shown in
Fig.~\ref{fig:supp_cross_category_gradient}(c), the category-gradient
norms vary substantially across insertion tasks.
The human category exhibits the largest average gradient magnitude and
the broadest variation, while the remaining categories generally
produce smaller updates with different levels of variability.

Under direct joint optimization, the shared update can be written as
\begin{equation}
    \mathbf{g}_{\mathrm{joint}}
    =
    \sum_{c}
    w_c
    \mathbf{g}_c,
    \label{eq:supp_joint_gradient}
\end{equation}
where \(w_c\) denotes the effective contribution of category \(c\).
Consequently, categories with larger gradient norms can dominate the
joint update.
When a large-magnitude category gradient is negatively aligned with
another category, it can further suppress the update required by the
latter.
Direct joint training is therefore affected by both directional
interference and category-dependent gradient-scale imbalance.

\paragraph{Implications for IOPD.}
The gradient analysis is consistent with the observed performance gap
between direct joint training and the proposed two-stage framework.
Direct Joint Training achieves a DINO-I score of \(0.6478\), a PSNR of
\(19.36\), and an LPIPS of \(0.1746\).
In comparison, the aggregated category-specific experts obtain
\(0.7187\), \(24.36\), and \(0.0777\), respectively, indicating that
learning each insertion category independently better preserves the
specialized behavior required by heterogeneous tasks.
After capability consolidation, the unified IOPD student achieves
\(0.7148\) DINO-I, \(24.40\) PSNR, and \(0.0795\) LPIPS, closely
approaching the category-specific experts while retaining a single model
for all insertion categories.

Together with the gradient observations in
Fig.~\ref{fig:supp_cross_category_gradient}, these results suggest that
the degradation of direct joint training is associated with weakly
aligned or recurrently conflicting category gradients, as well as
category-dependent differences in gradient magnitude.
Directly learning all insertion categories within one shared parameter
space can therefore suppress the specialized update directions required
by individual categories.

These observations motivate the central decoupling strategy of
InsertFuse.
Instead of requiring one model to acquire heterogeneous category
specializations directly from mixed-category supervision, we first train
each category expert in an independent parameter space.
This allows accessory, garment, human, animal, and general-object
insertion policies to develop without being weakened by incompatible
updates from other categories.

IOPD subsequently consolidates these learned policies using routed expert
supervision.
At each student-generated state, only the expert associated with the
current insertion category constructs the distillation target.
The unified student therefore learns from explicit category-specialized
velocity fields rather than attempting to discover all specialized
behaviors simultaneously from mixed-category ground-truth objectives.
The gradient results do not indicate that every category pair is
inherently conflicting; rather, they show that category compatibility is
heterogeneous and unstable under direct joint optimization.
This supports separating category-specific policy acquisition from
cross-category capability consolidation.
\end{document}